# ZeD-MAP: Bundle Adjustment Guided Zero-Shot Depth Maps for Real-Time Aerial Imaging

Selim Ahmet Iz[1,2], Francesco Nex[2], Norman Kerle[2], Henry Meissner[1], Ralf Berger[1]

[1]German Aerospace Center (DLR), Institute of Space Research, Berlin, Germany
(selim.iz, henry.meissner, ralf.berger)@dlr.de
[2]Faculty of Geo-Information Science and Earth Observation (ITC), University of Twente, Enschede, The Netherlands
(f.nex, n.kerle)@utwente.nl

## Abstract

Real-time depth reconstruction from ultra-high-resolution UAV imagery is essential for time-critical geospatial tasks such as disaster response, yet remains challenging due to wide-baseline parallax, large image sizes, low-texture or specular surfaces, occlusions, and strict computational constraints. Recent zero-shot diffusion models offer fast per-image dense predictions without task-specific retraining, and require fewer labelled datasets than transformer-based predictors while avoiding the rigid capture geometry requirement of classical multi-view stereo. However, their probabilistic inference prevents reliable metric accuracy and temporal consistency across sequential frames and overlapping tiles. We present ZeD-MAP, a cluster-level framework that converts a test-time diffusion depth model into a metrically consistent, SLAM-like mapping pipeline by integrating incremental cluster-based bundle adjustment (BA). Streamed UAV frames are grouped into overlapping clusters; periodic BA produces metrically consistent poses and sparse 3D tie-points, which are reprojected into selected frames and used as metric guidance for diffusion-based depth estimation. Validation on ground-marker flights captured at approximately 50 m altitude (GSD ≈ 0.85 cm/px, ~2,650 m² ground coverage per frame) with the DLR Modular Aerial Camera System (MACS) shows that our method achieves sub-meter accuracy, with approximately 0.87 m error in the horizontal (XY) plane and 0.12 m in the vertical (Z) direction, while maintaining per-image runtimes between 1.47 and 4.91 seconds. Results are subject to minor noise from manual point-cloud annotation. These findings show that BA-based metric guidance provides consistency comparable to classical photogrammetric methods while significantly accelerating processing, enabling real-time 3D map generation.

## 1. Introduction

The demand for rapid, operational aerial mapping and dense 3D reconstruction has grown sharply in domains where timely spatial intelligence can materially affect outcomes, notably in disaster response, search-and-rescue, and emergency infrastructure assessment. In such time-critical scenarios, near-instantaneous generation of georeferenced imagery and metrically accurate 3D products enables faster situational awareness, prioritization of resources, and safer operational planning for first responders. Operational toolchains developed at DLR for rapid mapping (Hein & Berger, 2018) and for first-responder support already target these use cases by minimizing latency between acquisition and map delivery and by integrating modular imaging hardware such as the DLR Modular Aerial Camera System (MACS).

Achieving real-time dense depth reconstruction from full-resolution UAV imagery is technically demanding for several reasons. Images often exceed 50 megapixels, flight altitudes and parallax conditions vary considerably, and disaster environments contain low-texture regions, repetitive structures, debris fields, and specular surfaces that degrade matching reliability. In this work, real-time refers to processing each incoming image pair within approximately two seconds, a requirement derived from typical UAV flight speeds (20m/s), overlap ratios (80%), and flight altitude (300m). Equally important, however, is metric generalizability: disaster sites differ drastically from one deployment to another, reducing the effectiveness of algorithms that rely heavily on pretraining or scene-specific assumptions. A method that achieves real-time speed but fails to maintain metric consistency across frames or across missions is insufficient for operational use.

Traditional dense reconstruction pipelines offer high geometric precision but are computationally intensive, especially at full resolution. Semi-Global Matching (SGM) (Hirschmuller, 2005) remains a widely used dense-stereo solution in photogrammetric workflows (and is part of operational DLR Institute of Space Research processing chains), because it produces dense depth estimates via robust, per-pixel disparity optimization and is comparatively resilient across diverse scenes. However, SGM's exhaustive pixel-wise matching and post-processing requirements imply high computational cost and make it poorly suited for strict sub-second/per-pair latency targets without substantial hardware acceleration.

Learned alternatives based on transformer architectures provide deterministic, per-image depth predictions and demonstrate strong performance on curated benchmarks, yet they depend heavily on large labelled datasets and powerful GPU hardware for both training and inference. Their generalizability to unseen disaster scenarios remains limited, particularly when imaging conditions deviate from the statistics of the training data. Diffusion-based depth estimation methods, in contrast, operate as probabilistic predictors and refine depth through iterative denoising. This iterative structure enhances robustness and cross-scene generalization, but it comes at the cost of slower inference and inconsistent scale across sequential frames, two shortcomings that hinder their direct use in time-critical UAV mapping.

These limitations motivate the development of hybrid strategies that combine the adaptability of diffusion models with geometric constraints capable of enforcing metric consistency. Such coupling promises a balanced trade-off between speed, accuracy, robustness, and generalization, enabling dense depth reconstruction that is both real-time and applicable to diverse, previously unseen environments without relying on expensive pretraining or external depth cues.

In this work, we introduce a Bundle Adjustment-guided diffusion depth prediction framework (Fig. 1) for large-scale depth reconstruction from full-resolution UAV imagery. Incoming frames are grouped into variable-size, overlap-aware clusters using estimated image footprints, which is the projected ground coverage of each frame, and bundle adjustment is run once per cluster to recover metrically consistent camera poses and sparse tie-points, which then condition a zero-shot diffusion model for dense depth estimation. When GNSS is unavailable, clusters fall back to fixed three-frame windows. This design yields the following contributions:

- To the best of our knowledge, this is the first real-time capable framework that delivers globally metric-consistent dense depth maps for large-scale aerial mapping using a test-time diffusion model.
- We introduce an overlap-aware cluster formation strategy in which image clusters are constructed either dynamically using GNSS-based overlap thresholds or fixed at three frames when GNSS is unavailable. This ensures that each cluster is compact yet sufficiently constrained for reliable metric initialization.
- We propose an efficient N-frame selection scheme that runs bundle adjustment only on the middle frame of each cluster and its immediate neighbours. This lightweight three-view BA produces a metrically consistent 3D anchor for the central frame, which is then used to condition the diffusion model once per cluster without compromising real-time performance.
- We provide an incremental, end-to-end pipeline that reads RGB images and provides dense depth map and true-ortho mosaics while also being capable of running with and without GNSS, thereby eliminating the need for external 3rd party solutions (e.g., COLMAP (Fisher et al., 2021)) or precomputed depth priors required by prior-guided zero-shot diffusion methods such as Marigold-DC (Viola et al., 2025) or Murre (Guo et al., 2025).

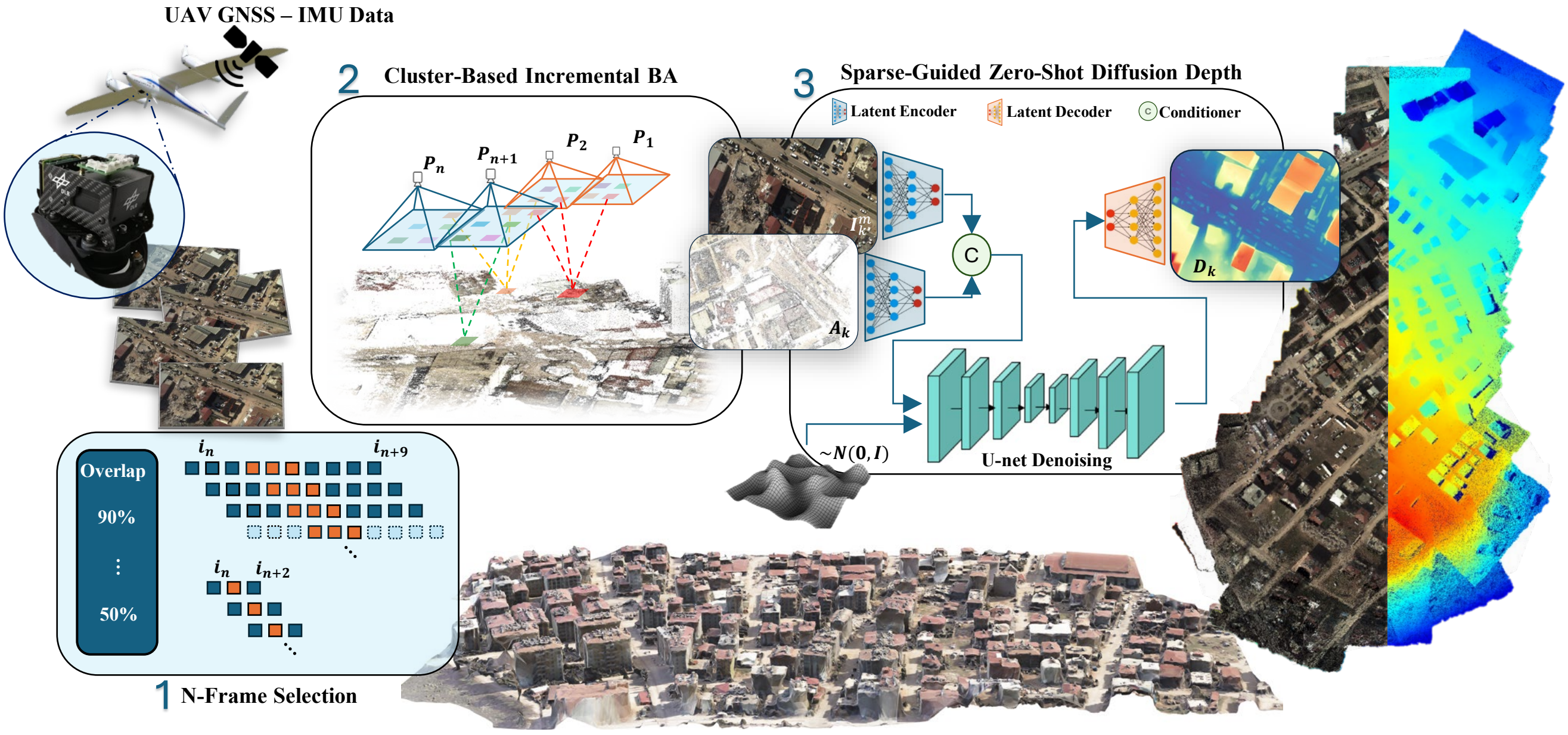

**Figure 1.** Integration of cluster-based bundle adjustment and diffusion model.

Together, these contributions form a scalable and operationally viable framework for real-time 3D reconstruction in challenging disaster environments, where both speed and metric reliability are indispensable.

The remainder of the paper places the proposed framework within related work (Section 2), details the cluster BA and diffusion-guidance modules (Section 3), and presents validation on MACS flights with ground markers and runtime analyses demonstrating the method's practical behaviour (Section 4). Lastly, Section 5 concludes the paper and discusses future research directions.

## 2. Related Work

Dense depth reconstruction for aerial imagery has a long history rooted in classical stereo and feature-based photogrammetry. Pixel-wise dense stereo methods, ranging from simple block-matching to the widely adopted SGM, explicitly exploit epipolar geometry to produce dense disparity maps with strong boundary preservation and robustness in low-texture areas; SGM's path-wise cost aggregation remains a practical standard for high-fidelity mapping on large-format sensors. At the same time, sparse feature pipelines based on repeatable keypoints (e.g., SIFT (Lowe, 2004) and ORB (Rublee et al., 2011)) followed by robust matching, triangulation and bundle adjustment deliver reliable camera geometry and metrically interpretable sparse 3D points.

Both families keep strong interpretability and geometric guarantees, but they scale poorly to ultra-high-resolution UAV frames: SGM requires large cost volumes and extensive pre/post-processing, while sparse SfM methods incur heavy matching and optimization costs as image counts and resolutions grow. Moreover, they are more error-prone structure in repetitive or low-texture regions compared to pixel-wise dense matching (Iz & Unel, 2023). These computational and scaling limitations are especially salient for on-the-fly, latency-sensitive mapping.

Building on these foundations, operational dense stereo systems follow a multi-stage workflow, precise rectification, cost aggregation, sub-pixel refinement, and geometric post-processing, to reach the accuracy required for mapping-grade DSMs. However, the cumulative memory and runtime demands of these stages rise steeply with image size and disparity range. Even optimized implementations, such as the SGM-based MACS pipeline at DLR, achieve their accuracy primarily in offline settings, as processing full-resolution pairs remains incompatible with strict real-time constraints without downsampling or specialized acceleration. This highlights the broader challenge: classical dense reconstruction offers strong geometric fidelity, but its computational overhead limits direct deployment in time-critical UAV scenarios.

To bridge the gap between geometric rigor and runtime constraints, the community has increasingly adopted learned components that accelerate or improve classical steps. Learned detectors and matchers - SuperPoint (DeTone et al., 2018), SuperGlue (Sarlin et al., 2020), LightGlue (Lindenberger et al., 2023) - increase match repeatability under adverse conditions and can reduce outliers, while learned modules that predict stereo regularization (SGM-Nets (Seki & Pollefeys, 2017)) or replace matching with learned correlation volumes (GC-Net (Tang et al., 2023) and Psmnet (Wang et al., 2022)) improve accuracy. These approaches often yield practical gains when selectively integrated (e.g., robust sparse matching followed by targeted dense stereo) yet they introduce new compute and memory budgets that favor GPU-equipped platforms and require careful design choices for UAV edge deployment (small backbones, adaptive inference).

Parallel to these hybrid approaches, deep learning methods have proliferated across monocular, stereo, and single-image zero-shot strategies. Monocular and SLAM-style models can run in real time and are increasingly trained or conditioned to provide scale-aware depth; however, many remain only up-to-scale unless calibrated or fused with metric sensors. Learned and foundation stereo networks produce metric depth by design, but they depend on adequate overlap and texture and typically require moderate GPU resources. Emerging feed-forward foundation and transformer-based models can generate rich per-frame geometry and even multi-view outputs, yet their reliance on very large pretraining corpora and heavier inference costs creates a tension between generalization and operational throughput for disaster response. For example, VGGT (Wang et al., 2025) demonstrates how per-frame feed-forward geometry can be integrated into a SLAM-like pipeline using lightweight temporal fusion to improve temporal stability and global consistency while keeping latency near single-pass levels. Similarly, MapAnything (Keetha et al., 2025) recently proposed a universal feed-forward transformer that regresses factored, metric 3D scene geometry (depth maps, local ray maps, camera poses, and a global scale) in one forward pass; this reduces downstream optimization but does not fully replace multi-view bundle adjustment. Encoder–decoder designs help mitigate some resolution trade-offs, yet they too trade computational load for recovery of fine detail. Overall, there remains a practical trade-off between metric accuracy, robustness, and inference efficiency, motivating hybrid systems that combine fast feed-forward predictions with selective geometric optimization for time-critical deployments.

Large-scale scene representations, Gaussian splatting and scaled NeRF variants, offer an alternative by condensing multi-view information into renderable scene models that can deliver very fast novel-view rendering and metric novel-view depth after substantial precomputation. Methods such as DroneSplat (Tang et al., 2025), SMERF (Duckworth et al., 2024) and other streaming/tiling strategies demonstrate that high visual fidelity and interactive rendering are achievable for city-scale scenes, but they typically require minutes to hours of preprocessing or distillation before real-time rendering is possible, which limits their applicability for immediate-response mapping without offline stages.

Recently, denoising diffusion priors have been repurposed for depth completion and single-image metric estimation in a zero-shot, test-time setting. Guided diffusion methods inject sparse metric cues (e.g., LiDAR samples or sparse SfM points) into the denoising process so that strong pretrained 2D priors are steered toward metrically meaningful outputs. Marigold-DC exemplifies the single-image usage: it requires an externally supplied, image-aligned sparse depth array (typically an .npy file) to guide per-image inference and does not recover depth directly from RGB nor enforce multi-view consistency. By contrast, Murre conditions diffusion on condensed 3D structure extracted offline via SfM toolchains (e.g., COLMAP) to produce multi-view consistent metric maps, but this design depends on offline preprocessing and is not intended for real-time operation. Although both strategies demonstrate impressive cross-scene generalization without per-scene retraining, their reliance on externally generated guidance, the iterative cost and stochasticity of diffusion inference, and sensitivity to the quantity and distribution of sparse cues constrain their applicability to time-critical UAV mapping. Addressing this guidance–consistency–cost trade-off is therefore central to adapting guided diffusion for real-time, large-scale aerial reconstruction.

Temporal fusion and hybrid online systems aim to further stabilize per-frame predictions: recurrent or stateful architectures (e.g., CUT3R (Wang et al., 2025)) and rollout/registration strategies (e.g., RollingDepth (Ke et al., 2025)) accumulate evidence across frames to reduce flicker and enforce temporal coherence, while SLAM-style wrappers around feed-forward Geometric Foundation Models (GFMs) (e.g., VGGT-SLAM (Maggio et al., 2025)) perform global alignment and loop closure to recover scale and long-term consistency. These approaches demonstrate that combining strong single-image priors with geometric post-processing can yield metric, temporally stable outputs, but they add optimization or registration costs that must be budgeted for real-time operation.

Taken together, classical methods provide reliable dense geometry but are costly at high-resolution UAV scales; learned and foundation models offer strong priors yet face scale ambiguity or high compute; diffusion-based zero-shot approaches can achieve metric depth with sparse tie-points but require careful cost management.

## 3. Methods

The proposed ZeD-MAP framework (Fig. 1) integrates a test-time diffusion depth estimator with an incremental cluster-based bundle adjustment module to obtain globally metric-consistent dense depth maps from ultra-high-resolution UAV image streams. Unlike classical multi-view stereo (MVS) pipelines, which require dense pairwise matching and global optimization, ZeD-MAP uses diffusion-based zero-shot monocular depth inference enriched with sparse metric tie-points derived from lightweight three-view BA. This design preserves the robustness and generalization capability of diffusion models while injecting the metric scale and cross-cluster consistency necessary for large-scale aerial mapping. The overall workflow operates sequentially during flight: incoming frames are assigned to overlap-aware clusters, sparse 3D points and camera poses are recovered through incremental local BA, the resulting 3D tie-points are reprojected into one representative frame per cluster, and finally the diffusion model produces a metrically guided dense depth map that inherits the anchor-based scale. Consecutive depth maps are finally fused into a global 3D reconstruction, including a true-ortho mosaic, a global depth map, and a dense 3D point cloud.

For all formulations, we denote an image by $I_k$ with intrinsic matrix $K$, camera rotation $R_k \in SO(3)$, translation $t_k \in R^3$, and a 3D point by $P_j \in R^3$. The projection operator is written as

$$u_{kj} = \pi(K(R_k P_j + t_k)), \pi(x, y, z) = \left(\frac{x}{z}, \frac{y}{z}\right) \quad (1)$$

Bundle adjustment minimizes robust reprojection error using a Huber penalty. Cluster indices are denoted $C_k$, and the middle image of the $k$-th cluster is denoted $I_k^m$. Depth maps predicted by the diffusion model are written $D_k$, and sparse metric tie-points for conditioning are denoted $A_k$. Finally, $d$, used in accuracy measurement, represents Euclidean distance.

### 3.1. Cluster-based Bundle Adjustment and Sparse Map Generation

The first component of ZeD-MAP is a sliding-window cluster-based BA scheme that produces sparse, metrically accurate 3D tie-points to guide diffusion-based depth estimation. For each incoming sequence of UAV images we form an overlap-aware cluster $C_k = \{I_t, \dots, I_{t+N-1}\}$, where $N$ is variable depending on GNSS availability (see Section 3.2). Within each cluster, we optimize camera poses $\{R_i, t_i\}$ and sparse 3D points $\{P_j\}$ by minimizing the classical BA energy

$$E = min_{\{R_i,t_i,P_j\}} \sum_{(i,j)\in O} \rho\left(\left|\pi\left(K\left(R_i\, P_j + t_i\right)\right) - u_{ij}\right|^2\right), \quad (2)$$

where $O$ denotes the set of valid observations and $\rho(\cdot)$ is the Huber cost. The optimization is solved using sparse Levenberg–Marquardt with Schur complement elimination, enabling real-time performance on local clusters even with 50+ MP imagery.

To enforce cross-cluster metric consistency, we adopt a pose-anchor protocol in which the last $M$ frames of the previous BA window remain fixed while optimizing the new cluster. These fixed frames serve as inter-cluster tie-points, preventing drift in both scale and orientation when moving from one cluster to the next. The strategy effectively emulates a SLAM-like keyframe graph while introducing no additional computational overhead beyond BA.

Feature observations for BA are obtained using a simplified matching strategy designed for real-time, wide-baseline UAV imagery. Our previous work (Iz et al., 2025) relied on dense patch tracking over predicted overlap regions; however, diffusion-based depth estimation requires sparse points to be well distributed over the entire frame. Patch-based tracking tends to concentrate matches in limited footprint-overlap regions, leaving large parts of the image without tie-points. Therefore, in ZeD-MAP we replace patch-tracking with RANSAC-filtered keypoint extraction, limit the total number of features per frame, and downsample images by 50% to reduce descriptor computation. Despite this simplification, the BA remains sufficiently accurate to match image and the generated 3D point cloud because the three-view cluster selection method (Section 3.2) ensures that the middle frame receives dense and well-spread 3D points.

The final output of each cluster BA is a metrically correct set of camera poses and a sparse 3D reconstruction anchored to the UAV's GNSS frame when available. These sparse points are then projected into the cluster's representative frame $I_k^m$, producing a depth-aligned anchor map $A_k$ with associated per-point uncertainty estimates that guide the subsequent diffusion process.

### 3.2. N-Frame Selection

In UAV mapping, successful reconstruction of a well-distributed 3D point cloud requires sufficient overlap between at least two images. Standard two-view geometry produces 3D points only in the overlap region, leaving uncovered areas without anchor support. Therefore, clusters in ZeD-MAP must contain at least three sequential images to ensure that the middle frame receives near-full coverage of 3D points, provided that adjacent images share at least 50% overlap.

We distinguish two scenarios depending on whether GNSS measurements are available. If GNSS is absent, clusters are fixed to triples of sequential images $\{I_{k-1}, I_k, I_{k+1}\}$. BA is applied to these three frames, and the middle frame $I_k$ becomes the representative cluster frame. The minimal cluster size of three ensures that 3D points are generated across the full footprint of $I_k$, assuming the standard UAV acquisition geometry with at least 50% forward overlap.

When GNSS is available, we adopt a dynamic cluster formation approach. Starting from the first image of a cluster, incoming frames are appended until the estimated overlap relative to the first frame drops below a preset threshold (10% in our implementation). For example, with 90% forward overlap in the image sequence, a terrain features visible at the trailing edge of $I_1$ will continue to appear in $I_2, I_3, \dots$ until eventually disappearing beyond $I_9$. Without reduction, BA over all nine frames would increase runtime significantly due to the large number of optimization parameters. However, the marginal gain in 3D point density diminishes once baseline diversity is sufficiently expressed.

To balance coverage and computational load, ZeD-MAP introduces an efficient N-frame selection strategy. Given a dynamically formed cluster with potentially many images $\{I_1, \dots, I_L\}$, we perform BA only on a subset of three views: the middle frame $I^m$ and its immediate neighbors $I^{(m-1)}$ and $I^{(m+1)}$. Formally, if the cluster spans frames $\{t, t+1, \dots, t+L-1\}$, the BA window is $\{I_{t+m-1}, I_{t+m}, I_{t+m+1}\}$, where $m = \lfloor L/2 \rfloor$. This three-view BA is sufficient to generate a full-coverage 3D anchor set for $I^m$ because the frame is centrally located within the overlap range, implying maximal support from neighboring views. The reduced BA complexity, from dozens of frames to only three, preserves real-time performance while maintaining metrically consistent reconstruction.

Once BA converges, we store only the results for the middle frame of each cluster: its optimized pose, its sparse depth tie-points $A_k$, and the frame itself. Subsequent clusters inherit metric consistency through the fixed-frame mechanism described in Section 3.1. As the UAV stream progresses, the system accumulates a sequence of representative frames, each accompanied by a reliable, dense 3D anchor set that fully covers the frame, enabling consistent and scalable guidance for the diffusion model.

### 3.3. Guided Diffusion Model

The final component of ZeD-MAP is a guided diffusion module that converts the BA-derived sparse tie-points into a metrically accurate, dense depth map for each representative cluster frame. Conventional diffusion-based monocular depth models, such as Marigold-DC or Murre, provide strong generalization across diverse scenes but lack absolute scale and produce depth predictions that may vary across overlapping frames. Among available methods, Murre exhibits superior performance for aerial imagery; its conditioning mechanism based on sparse SfM projections aligns naturally with our BA-generated anchor points. We therefore adopt Murre as the backbone and adapt it to operate incrementally within our real-time pipeline.

In each cluster, the sparse anchor set $A_k$ is projected into the representative frame and rasterized into a depth-aligned grid matching the image resolution. Missing pixels are left unfilled, preserving the sparsity pattern arising from geometric visibility. Camera intrinsics and the optimized pose of $I_k^m$ are then provided to the diffusion model as additional conditioning signals. The diffusion process operates in latent space: a noisy latent depth map is iteratively denoised using both the RGB content and the sparse metric tie-points. Through this conditioning, the depth prediction

$$D_k = Diffusion(I_k^m, A_k, K, R_k, t_k) \quad (3)$$

inherits the scale enforced by BA while retaining the high-frequency structural accuracy typical of diffusion-based monocular depth estimators.

Unlike the original Murre implementation, which performs full-scene reconstruction offline using global COLMAP points, our adaptation treats Murre as a grey-box module operating sequentially on cluster frames. After predicting the dense map $D_k$, we fuse it incrementally into a global reconstruction using truncated signed distance functions (TSDF). The TSDF volume is updated online with each new depth map, progressively accumulating depth observations from all representative cluster frames. This enables real-time visualization of the evolving dense point cloud.

Finally, the fused TSDF is converted into three global products:

- a globally consistent dense 3D point cloud,
- a global depth map reconstructed from the integrated TSDF, and
- a true-orthomosaic obtained by projecting radiometric information onto an orthorectified grid.

These outputs reflect the metric accuracy introduced by BA and the dense scene geometry recovered by diffusion, combining the strengths of probabilistic monocular inference with the determinism of geometric optimization.

### 3.4. Validation and Benchmarking

We evaluate the proposed BA-guided diffusion framework through a three-stage validation protocol designed to assess (i) cross-frame depth consistency, (ii) metric accuracy against recent large-scale monocular depth estimation baselines, and (iii) operational scalability on real disaster-response imagery.

**3.4.1. Cross-Frame Consistency of Guided Diffusion:** We first examine the effect of BA-generated sparse tie-points on the temporal and geometric consistency of the diffusion model. To this end, we compare unguided zero-shot diffusion predictions with the proposed guided formulation over sequential UAV frames. The objective is to qualitatively assess whether the cluster-level metric cues stabilize scale, suppress stochastic frame-to-frame fluctuations inherent to diffusion sampling, and better preserve structural continuity across large parallax transitions. This analysis highlights the specific contribution of cluster-based guidance independent of absolute metric accuracy.

**3.4.2. Accuracy on Ground-Marker Dataset:** To quantify metric performance we evaluate our method on a controlled dataset consisting of 22 high-resolution aerial images (8064 × 4536 px) acquired at an approximate flight altitude of 50 m. Within the scene, ground markers (Fig. 2) are placed at varying altitudes and planimetric separations, providing well-defined geometric baselines for assessing reconstruction accuracy. The manually selected 3D marker coordinates extracted from the reconstructed point cloud serve as reference measurements, allowing us to compute inter-marker relative errors on both the horizontal plane (X-Y) and the vertical axis (Z). Inter-marker relative errors are computed by comparing the distances between reconstructed marker points and their known ground-truth separations. For each marker pair, the relative error is defined as:

$$E_{rel} = \frac{|d_{measured} - d_{gt}|}{d_{gt}} x100\% \tag{4}$$

These evaluations directly assess the framework's ability to maintain metric scale and spatial coherence in large-baseline aerial imagery.

For comparison we benchmark our method against several representative approaches from the literature:

- VGGT and MapAnything: two state-of-the-art feed-forward geometric predictors with strong cross-scene generalization stemming from large-scale pretraining. Although they do not perform SLAM-style optimization or enforce multi-view consistency, they are suitable baselines for fast, monocular metric depth estimation.
- COLMAP: included as a classical offline baseline with high geometric fidelity but no real-time capability.

These baselines were selected because they represent the most capable publicly available methods for fast monocular or pseudo-monocular metric depth estimation, benefit from extensive pretraining, and reflect the upper bounds of what can be achieved without SLAM-level geometric optimization.

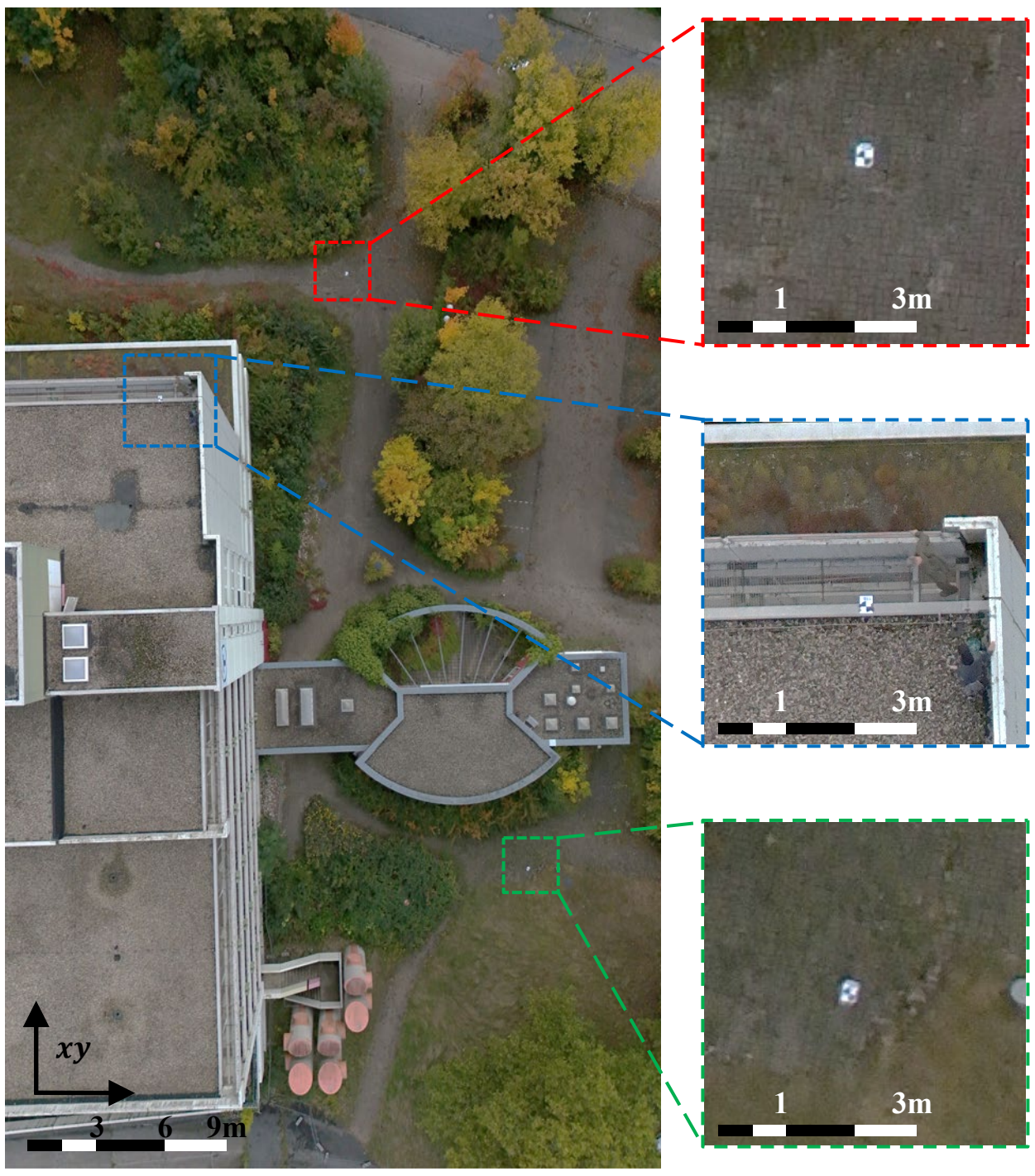

**Figure 2 .** Ground-marker locations on test scene.

**3.4.3. Large-Scale Evaluation on Earthquake Imagery:** We further assess the proposed framework on a large-scale, multi-strip UAV dataset acquired over an earthquake-affected region. The dataset consists of 60 aerial images at a resolution of 7920 × 6004 pixels, captured from an approximate flight altitude of 325 m. This dataset does not contain ground markers; instead, it serves to evaluate qualitative reconstruction fidelity, multi-strip consistency, and operational scalability under realistic disaster-mapping conditions. The scene includes severe structural damage, debris, large depth discontinuities, and low-texture surfaces, making it a representative stress test for real-time aerial mapping.

Instead of focusing on per-method benchmarking as in the controlled marker experiment, this large-scale scenario is used to examine how the proposed system behaves under extended flight paths, heterogeneous scene geometry, and cross-strip interactions. The emphasis is placed on the stability of the reconstruction over hundreds of meters, the ability to maintain coherent depth across wide-baseline strip intersections, and the resilience of the N-frame scheduling and cluster-level BA anchoring under operational load. This evaluation highlights how our method preserves multi-view consistency and suppresses drift accumulation in settings where real-time feed-forward predictors typically show fluctuating scale, depth discontinuities, or frame-to-frame inconsistency.

In addition to qualitative assessment, we compute several internal DSM-derived quality indicators that do not require ground-truth data, derived from the ortho-rasterised height map $H \in R^{N_r \times M_r}$, where $N_r$ and $M_r$ are the raster grid dimensions and each cell $H_p$ records the maximum observed elevation within its ground footprint. $V = \{p: H_p \neq NaN\}$ denotes the set of cells with at least one valid fused depth observation, and $|V|$ its cardinality. Coverage, which is defined as $Coverage = |V|/(N_r \cdot M_r)$, reflecting the spatial completeness of the reconstruction. Global standard deviation,

$$\sigma_{global} = \sqrt{\frac{1}{|V|}\sum_{p \in V}\left(H_p - \bar{H}\right)^2} \quad (5)$$

captures scene-level elevation dispersion where $\bar{H} = \frac{1}{|V|}\sum_{p \in V} H_p$ is the mean elevation over all valid cells. As a robust alternative resilient to outliers and blunders endemic to disaster imagery, the Normalised Median Absolute Deviation is defined as:

$$NMAD = 1.4826 \cdot median_{p \in V}|H_p - \tilde{H}| \quad (6)$$

where $\tilde{H}$ is the median elevation over $V$ and the 1.4826 scaling factor renders NMAD asymptotically equivalent to the standard deviation under Gaussian noise while maintaining a high breakdown point. To capture small-scale surface roughness, mean local standard deviation $\bar{\sigma}_{local}$ is computed over a $k \times k$ sliding window (default $k = 3$):

$$\bar{\sigma}_{local} = \frac{1}{|V'|}\sum_{p \in V'} \sigma_{local}(p) \quad (7)$$

where $V' = \{p \in V: |N(p)| > 1\}$ is the subset of valid cells with more than one valid neighbour in the window, $N(p)$ denotes that neighbourhood, and $\sigma_{local}(p)$ is the standard deviation of $H$ over $N(p)$. This makes $\bar{\sigma}_{local}$ sensitive to fine-grained depth noise and fusion artefacts rather than macro-scale geometry. Together, these indicators capture large-scale height stability ($\sigma_{global}$), robustness to outliers ($NMAD$), fine-scale surface smoothness ($\bar{\sigma}_{local}$), and spatial completeness ($Coverage$). Lower $NMAD$ and ($\bar{\sigma}_{local}$) indicate reduced noise and improved surface coherence, while high coverage reflects consistent depth prediction across the scene.

## 4. Results and Discussion

This section presents the experimental findings derived from the three-stage aforementioned validation protocol. We report qualitative evaluations of depth consistency across sequential frames, quantitative accuracy based on a ground-marker dataset, and large-scale performance on earthquake imagery. Unless stated otherwise, all diffusion-based depth predictions were executed in Python on an NVIDIA RTX A6000 GPU, whereas bundle adjustment and geometric preprocessing steps were run in MATLAB on a workstation equipped with a 13th-generation Intel Core i7 (20-core) CPU and 32 GB RAM.

### 4.1. Cross-Frame Depth Consistency

We first assess the temporal consistency benefits of BA-guided diffusion by comparing sequential frame predictions produced by several zero-shot or pretrained monocular depth estimators. The evaluated methods include RGB input only, DINOv2, DepthPro, Depth Anything v2, Marigold, and the guided diffusion model used in our pipeline.

Qualitative inspection (Fig. 3) shows that unguided diffusion exhibits noticeable frame-to-frame fluctuations in scale, depth discontinuities, and stochastic surface perturbations, especially over regions with weak texture or large parallax. Pretrained monocular models (DepthPro, Depth Anything v2, Marigold) provide more stable per-frame predictions, but still lack inter-frame consistency, as illustrated by the arrows indicating discrepancies between consecutive frames. By contrast, cluster-level BA guidance substantially reduces temporal drift, stabilizes depth across overlapping views, and preserves structural continuity in building facades, debris fields, and terrain undulations. These observations confirm that even sparse BA tie-points, injected once per cluster, are sufficient to suppress diffusion-induced stochasticity and enforce cross-frame geometric coherence.

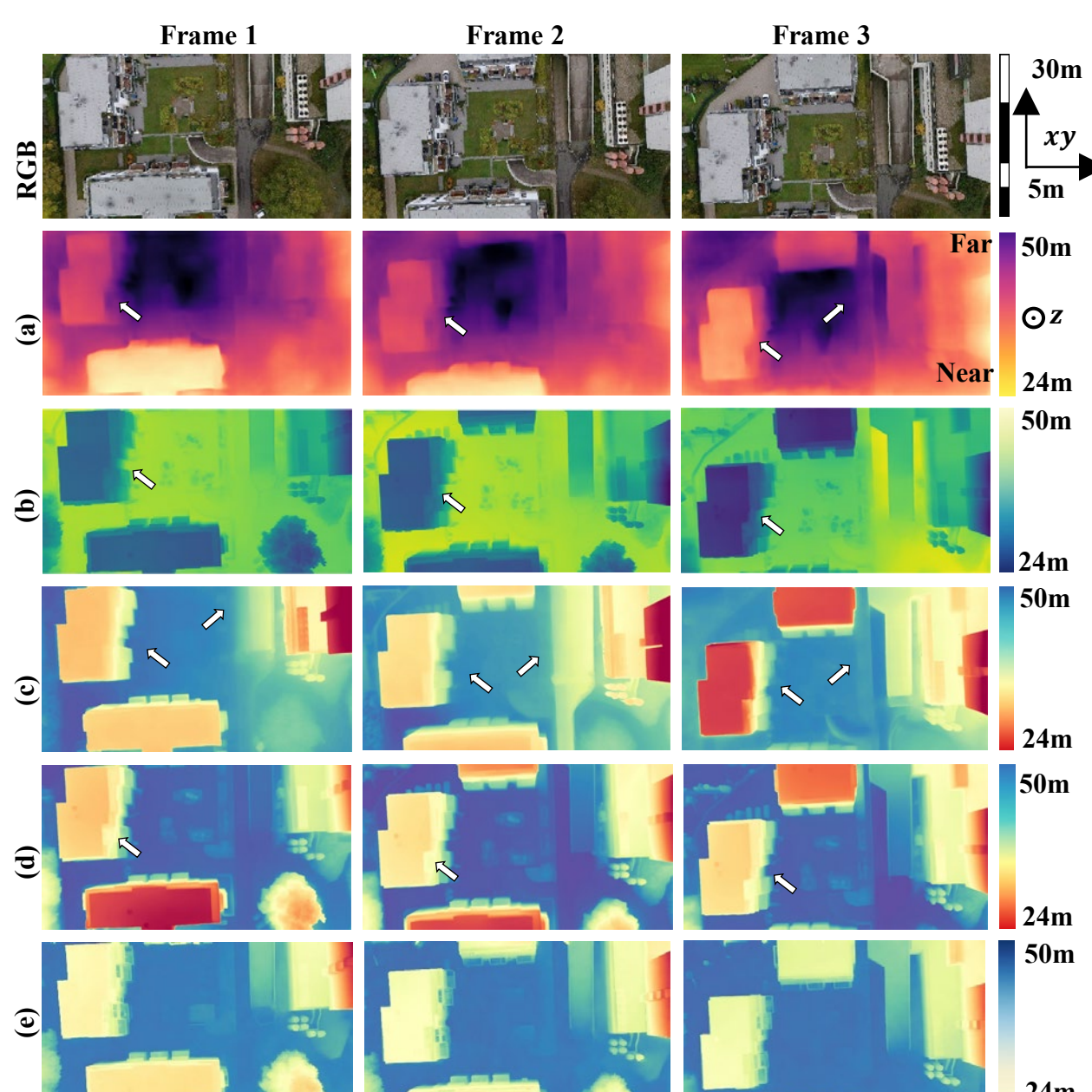

**Figure 3.** Comparison of depth maps generated by: (a) DINOv2, (b) DepthPro, (c) Marigold, (d) DepthAnything v2, and (e) ZeD-Map. Small arrows indicate inconsistencies between sequential depth maps. Depth maps shown in their native color scales to avoid artifacts and detail loss from normalization, with color bars indicating specific depth values.

### 4.2 Accuracy on Ground-Marker Dataset

We next evaluate absolute metric fidelity using a controlled dataset containing ground markers with known separations in both planimetric (XY) and vertical (Z) directions. Marker coordinates were manually selected from the reconstructed point clouds generated by each method. For every marker pair, we compute the inter-marker relative error Eq (4), then multiplied with actual distances between ground-markers to get metric values.

| | $e_{xy}$ $(m)$ | $e_z$ $(m)$ | Run-Time |
|---|---|---|---|
| COLMAP | 0.855 | 0.057 | 45.22s /image |
| VGGT | 1.355 | 0.734 | 1.99s /image |
| MapAnything | 1.109 | 0.316 | 1.47s /image |
| ZeD-Map | 0.867 | 0.123 | (1.47, 4.91)s* /image |

**Table 1:** Absolute positioning errors (meters) and runtime comparison. Errors based on mean separations of three ground-truth markers: 40 m (XY) and 26 m (Z) *ZeD-Map runtime is reported per cluster and normalized to per-image runtime assuming clusters of 3–10 image pairs.

Planar (XY) and vertical (Z) accuracy are then derived from the absolute relative deviation. Table 1 summarizes quantitative results comparing COLMAP, VGGT, MapAnything, and the proposed BA-guided diffusion approach, including both metric accuracy and runtime per image. As expected, COLMAP achieves strong geometric fidelity but at the cost of long runtimes and a non–real-time processing pipeline. VGGT and MapAnything yield competitive single-image scale estimates thanks to large-scale pretraining but show weaker relative accuracy, particularly in height differences.

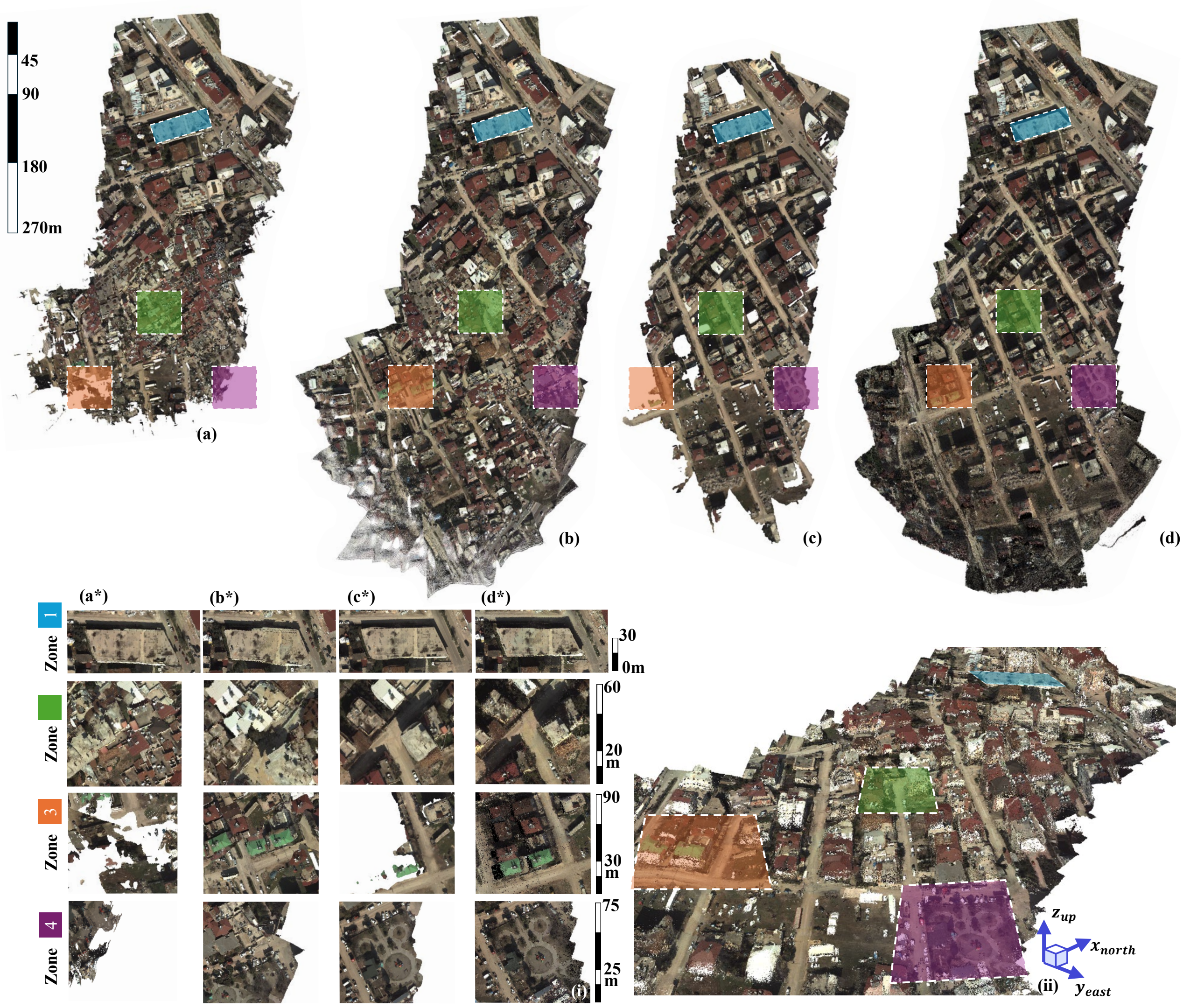

**Figure 4.** Point clouds generated by (a) VGGT, (b) Map Anything, (c) COLMAP, (d) proposed method. (i) Enlarged views of target regions from corresponding point clouds (a* to d*). (ii) The perspective view of selected regions in the point cloud generated by the proposed method.

The proposed method achieves second highest overall metric consistency, with 0.867 m error in XY and 0.123 m in Z, corresponding to relative errors below 2.2% for XY and below 1% for Z separations. The per-cluster processing time of our approach is 14.72 s, where each cluster has 3-10 images depending on the overlap ratio, resulting in an efficient runtime of 1.47-4.91s per image pair. This time includes the diffusion-based depth reconstruction stage accelerated on GPU (10 s per image, using 10 iterations, ensemble size 5, and width downsampled to 1024 px). During bundle adjustment, RANSAC is applied and the maximum number of detected features is limited to 4000 to balance accuracy and runtime. These results indicate that lightweight cluster BA, combined with constrained feature selection, successfully provides a reliable metric foundation for diffusion-based dense depth reconstruction while maintaining runtimes compatible with real-time UAV mapping.

### 4.3 Large-Scale Earthquake Dataset

Finally, we validate the framework on a 60-image, two-strip UAV dataset acquired by the DLR MACS during the 2023 Türkiye earthquake mission. Each image has a resolution of 7920×6004 px. This dataset, lacking ground markers, is used to assess multi-strip consistency, qualitative reconstruction fidelity, and operational runtime. Diffusion inference again runs on an RTX A6000 GPU, while BA executes on CPU as noted above.

Qualitative results (Fig. 4) show that VGGT and MapAnything lose geometric stability when a photogrammetric block with multiple strips is considered. As seen in the side-view point clouds, the drift is not limited to XY misalignment: both VGGT and MapAnything also exhibit noticeable Z-axis inconsistencies, with depth gradually diverging across frames. ZeD-Map maintains substantially better inter-strip alignment, though a small height inflation appears at the rotation zone. This artifact stems from the absence of a known yaw angle during this test and the diffusion model's attempt to remain within the user-specified maximum height range.

Overall, ZeD-Map and COLMAP are the only methods that maintain consistent global geometry across both strips. Therefore, these two approaches are compared quantitatively in Table 2. The results indicate that COLMAP achieves lower global deviations, as expected from a full offline SfM–MVS pipeline, while ZeD-Map reaches comparable local precision with higher coverage and significantly faster processing, making it suitable for real-time mapping.

| | COLMAP | ZeD-Map |
|---|---|---|
| Global std-dev (m) | 0.23 | 0.46 |
| Mean local std dev (m) | 0.023 | 0.024 |
| NMAD (m) | 0.26 | 0.55 |
| Coverage | 0.51 | 0.53 |

**Table 2:** Quantitative comparison of COLMAP and ZeD-Map.

The results in Table 2 show that COLMAP attains the lowest global deviation due to its full offline SfM–MVS optimization pipeline. In contrast, the proposed method achieves nearly identical local precision to COLMAP (0.023 vs. 0.024 m) while delivering slightly higher coverage and operating at a fraction of the computational cost. Although it does not reach COLMAP's level of global smoothness, the proposed method preserves photogrammetric-grade local consistency and remains far more suitable for time-critical, real-time UAV mapping scenarios.

## 5. Conclusion

In this work, we introduced ZeD-MAP, a near–real-time, metrically consistent depth-mapping framework that tie-points zero-shot diffusion predictions with lightweight, cluster-based bundle adjustment. By combining overlap-aware cluster formation, an efficient three-view BA strategy, and sparse metric conditioning of a test-time diffusion model, our approach delivers globally coherent dense depth maps from UAV imagery without relying on traditional SfM pipelines or extensive training data. Experiments on MACS aerial datasets demonstrate that BA-guided diffusion can substantially improve the metric stability of zero-shot depth estimates, achieving over 90% relative inter-marker accuracy while remaining suitable for real-time operation.

Despite these advantages, ZeD-MAP is still constrained by diffusion's computational overhead and the challenges of scaling to very large image resolutions. Future work will focus on further reducing inference latency, enabling processing of larger frames, and exploring tighter integration between diffusion models and SLAM-style optimization to improve robustness under extreme viewpoints, low-texture surfaces, and rapid motion.